\documentclass[11pt]{article}

\usepackage[final]{acl}

\usepackage{times}
\usepackage{latexsym}
\usepackage[T1]{fontenc}
\usepackage[utf8]{inputenc}
\usepackage{microtype}
\usepackage{inconsolata}
\usepackage{graphicx}
\usepackage{amsmath}
\usepackage{amssymb}
\usepackage{booktabs}
\usepackage{multirow}
\usepackage{hyperref}

\title{Influcoder: Distilling Decoders' Gradient Influence Rankings \\
into an Encoder for Data Attribution}

\author{
Dimitri Kachler, Damien Sileo, Pascal Denis \\
Centre Inria de l'Université de Lille, CRIStAL, Université de Lille \\
\{dimitri.kachler, damien.sileo, pascal.denis\}@inria.fr
}
\begin{document}
\maketitle

\begin{abstract}
Data attribution traces model behavior back to individual training examples, with applications such as identifying training data that induces toxic responses or tracing a model's factual claims back to the examples that support them. Gradient-based influence methods provide reliable attribution performance, yet are impractical at the scale of LLM training datasets due to their high computational costs. We introduce \textsc{Influcoder}, which distills the LoRA-gradient cosine influence rankings of a target decoder into a compact encoder, reducing attribution at inference time to an embedding cosine similarity. It scores training examples 35x faster than gradient-based influence computed on the target model and 18x faster than influence computed on the smallest comparable proxy model. Our method remains competitive with gradient-based methods even when considering the setup cost, as evaluated on two downstream tasks of toxicity filtering and factual attribution.
\end{abstract}

\section{Introduction}

Data attribution asks how individual training samples shape model behavior \citep{Deng2025}. In practice, data attribution is used to curate data to improve a target capability such as reasoning \citep{xia2024less}, to filter out examples that cause harmful responses, or to attribute claims to the underlying examples that support them \citep{2025datelm}. Attribution requires scoring each sample in the training \textit{pool} on its relevance to the samples in the \textit{query} set. For instance, the query set could include examples of toxic question-answer pairs. The computational costs therefore grow with the size of the pool and query sets, which becomes prohibitive at the scale of modern LLM training  datasets. 

Influence functions \citep{InfluenceDeepNN} are used for a variety of data attribution tasks, with a common motivation being \textit{data selection} to find the optimal set of training data. Pre-existing work \citep{xia2024less} has demonstrated influence-based selection for SFT on LLMs. On the other hand, \citet{ivisondata2025} demonstrate that an activation-based method can outperform influence-based selection at larger scales. \citet{li-etal-2025-influence} also report consistent performance issues, which they partly attribute to inverse Hessian approximation errors. On the other hand, influence methods consistently outperform against baselines on applications such as \textit{toxicity filtering} and \textit{factual attribution} \citep{2025datelm}, which motivates these two attribution tasks as the regimes we target. While performant, influence methods require per-example gradients on the entire pool and query sets and thus limits their applicability on large datasets.

\paragraph{Contribution}

We introduce \textsc{Influcoder}, which distills the LoRA-gradient influence rankings of a target decoder into a compact encoder to provide influence-based attribution at scale. We increase efficiency on both inference speed and storage space, processing examples 18$\times$ faster than the quickest comparable proxy. Our code is publicly available at: \href{https://github.com/dkachlerinria/influcoder}{https://github.com/dkachlerinria/influcoder}

\begin{figure*}[t]
    \centering
  \includegraphics[width=0.78\linewidth]{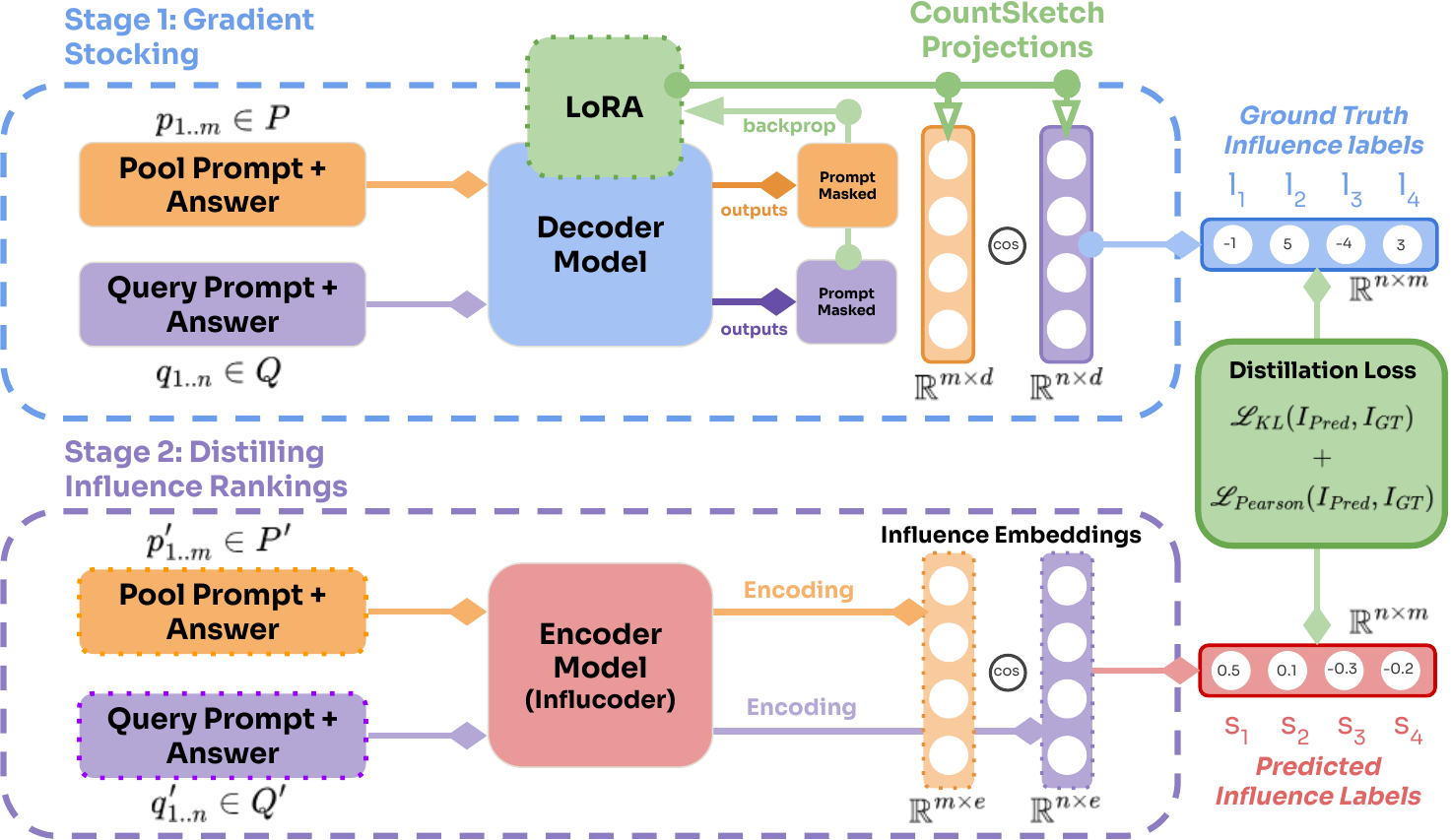} \hfill
  \caption {Schema for the training pipeline of \textsc{Influcoder}. Stage 1: full pool and query samples are passed to the target decoder model, then we compute LoRA gradients on the responses only. Finally, we project the gradients with CountSketch and compute ground truth cosine similarity influences. Stage 2: full pool and query samples are passed through the encoder to produce influence embeddings, cosine similarity is taken to generate influence predictions.}
\label{schema}
\end{figure*}

\section{Related Work}
Data Attribution methods for LLMs fall into one of three families, distinguished by the space in which they perform attribution.

\paragraph{Representation methods} 
Representation-based methods function in training-agnostic feature spaces, meaning they tend to work without setup and are generally cheap to compute. Examples include TF-IDF, dual-encoders for semantic similarity \citep{ni-etal-2022-large}, and RDS+ \citep{ivisondata2025}, which uses the model's activation space.

\paragraph{Gradient Influence methods}

\citet{Grosse2023StudyingLL} scale influence functions to LLMs using dense, unprojected gradients, whereas projection-based methods compress gradients so that they can be stored and reused. TRAK \citep{park2023trak} randomly projects the gradients, which approximately preserves inner products \citep{Johnson1984ExtensionsOL}.

TrackStar \citep{chang2025scalable} and LoGra \citep{choe2025what} instead apply two-sided projections to each layer's gradient matrix, which LoGra implements with small auxiliary layers on top of linear layers to keep a low memory footprint. LoRIF \citep{li2026loriflowrankinfluencefunctions} further improves on storage and query time by leveraging the rank-c factors of projected gradients. Methods such as LESS \citep{xia2024less} leverage LoRA \citep{hu2022lora} to calculate influence from gradients on the much smaller adapters. DataInf \citep{kwon2024datainf} and RRInf \citep{tu-etal-2025-rrinf} provide efficient estimations for the inverse Hessian for LoRA based influence.

\paragraph{Distillation methods}
Distillation based methods amortize attribution by paying a one-time initial setup cost to train a smaller surrogate model, the small surrogate model is then leveraged to achieve quicker inference speeds. IProX \citep{chen2026influencepreserving} constructs gradient-aligned proxies to estimate the full gradient influences for a target model. AirRep \citep{Sun2025AirRep}  is the closest method to ours, it also trains an encoder whose embeddings' dot products predict training influence. However, the supervision signal is fundamentally different, AirRep distills empirical training losses collected from thousands of small LLM training runs, whereas we distill single-checkpoint gradient influences of the target decoder. Furthermore, its embeddings are pooled over candidate sets with attention to capture group effects, whereas \textsc{Influcoder} treats samples independently which is more fitting for attribution.

\section{Methodology}

\paragraph{LoRA Cosine Influence}

Classic influence formulations \citep{InfluenceDeepNN} weight the gradient
inner product by an inverse Hessian to account for loss curvature. Estimating
that term can be unstable, thus we work from the first-order formulation of \citet{TracIn}. Following LESS
\citep{xia2024less} we further restrict the gradients to the LoRA
\citep{hu2022lora} adapter parameters $\theta_L$ and replace the dot product
with cosine similarity, which reads gradient direction rather than magnitude
and so avoids over-weighting samples with large gradient norms,
\begin{equation}
Inf(p,q) =  cos(\nabla \ell(\theta_{L}, p) , \nabla \ell(\theta_{L}, q))
\end{equation}

\subsection{Collecting Gradient Influence Labels}

Let $\mathcal{P} = \{p_i\}_{i=1}^{N_P}$ and $\mathcal{Q} = \{q_j\}_{j=1}^{N_Q}$ 
denote the pool and query sets, respectively. We collect the ground truth gradients from the pool and query set only through the LoRA parameters, see Figure~\ref{schema} and also mask the prompt, since response-only gradients are more informative. To save on storage, we randomly project the gradients to a smaller dimension using CountSketch \citep{CountSketch} projection. In Appendix~\ref{Appendix-A1} we show how the projection dimension conserves the influence rankings on different projection sizes. We exclude the LM head and the vocabulary embeddings, as they are noisy and hinder the training.

\subsection{Ranking-Based Influence Distillation}

At the start of the training, we pre-compute all of the cosine-similarity influences between the pool samples $p_i$ and query samples $q_j$ to easily access the ground truth influences during training from the influence matrix, $I_{GT} \in \mathbb{R}^{N_P \times N_Q}$. On each batch we sample $m$ pool and $n$ query samples and pass them through the encoder to get the batch of influence embeddings $E^p \in \mathbb{R}^{m \times e} $ and $E^q \in \mathbb{R}^{n \times e}$, on which we take the cosine similarities to get the predicted influence, $[I_{P}]_{ij} = \cos(E^p_i,\, E^q_j)$. We then fetch the submatrix $I_{GT}^{\mathrm{b}} \in \mathbb{R}^{m \times n}$
from $I_{GT}$ using the sampled indices.

 \paragraph{Ranking Losses}
 
 While we would like to optimize on the Spearman metric directly, it is a non-differentiable function. However, we can use a surrogate ranking metric through the Pearson correlation, which is differentiable, we simply flatten the influence matrices and take one minus the Pearson correlation,

\begin{equation}
    \mathcal{L}_{\mathrm{Pearson}} 
  = 1 - 
  \frac{\mathrm{cov}\!\left(\mathrm{flat}(I_{P}),\, \mathrm{flat}(I_{GT}^{\mathrm{b}})\right)}
       {\mathrm{std}(I_{P})\, \mathrm{std}(I_{GT}^{\mathrm{b}})}
\end{equation}

The advantage to the Pearson correlation is that it aligns the entire batch at once. As a consequence it also means that it does not particularly optimize for the top-k most influential samples. Thus, we add a softmax KL divergence loss \citep{Hinton2015DistillingTK}, 
\begin{equation}
  \mathcal{L}_{\mathrm{KL}} 
  = \frac{1}{n} \sum_{j=1}^{n} 
          \mathrm{KL}\!\big(
      \sigma_\tau\!\big([I_{GT}^{\mathrm{b}}]_{:,j}\big) 
      \,\|\, 
      \sigma_\tau\!\big([I_{P}]_{:,j}\big)
    \big)
\end{equation}
For each query $j$ we calculate the column 
$[I_{P}]_{:,j} \in \mathbb{R}^{m}$ giving influence scores over pool candidates and 
define the tempered softmax with temperature $\tau$ as
$\sigma_\tau(z)_i = \exp(z_i/\tau) / \sum_{k=1}^{m} \exp(z_k/\tau)$. The per-query KL loss is then averaged over the $n$ queries in the batch. Finally, we combine both losses linearly with an $\alpha \in[0,1]$ term,
\begin{equation}
    \mathcal{L}_{\mathrm{Total}} =
    \alpha\, \mathcal{L}_{\mathrm{Pearson}}
    + (1-\alpha)\, \mathcal{L}_{\mathrm{KL}}
\end{equation}

\paragraph{High-influence sampling}
Depending on the data distributions, many candidate samples may have low influence for all query samples. Thus, there might be limited signal in separating high-influence samples from the rest of the pool distribution. We therefore fill a \textit{high-influence ratio} of the pool slots with samples that carry high individual influence on the query samples. Each query in the batch contributes its top-ranked pool
samples under $I_{GT}$, and the union of these is subsampled
uniformly to fill the high-influence slots. The remaining
slots are drawn uniformly from the rest of the pool.

\section{Experimental Setup}

\begin{figure*}[t]
  \includegraphics[width=0.97\linewidth]{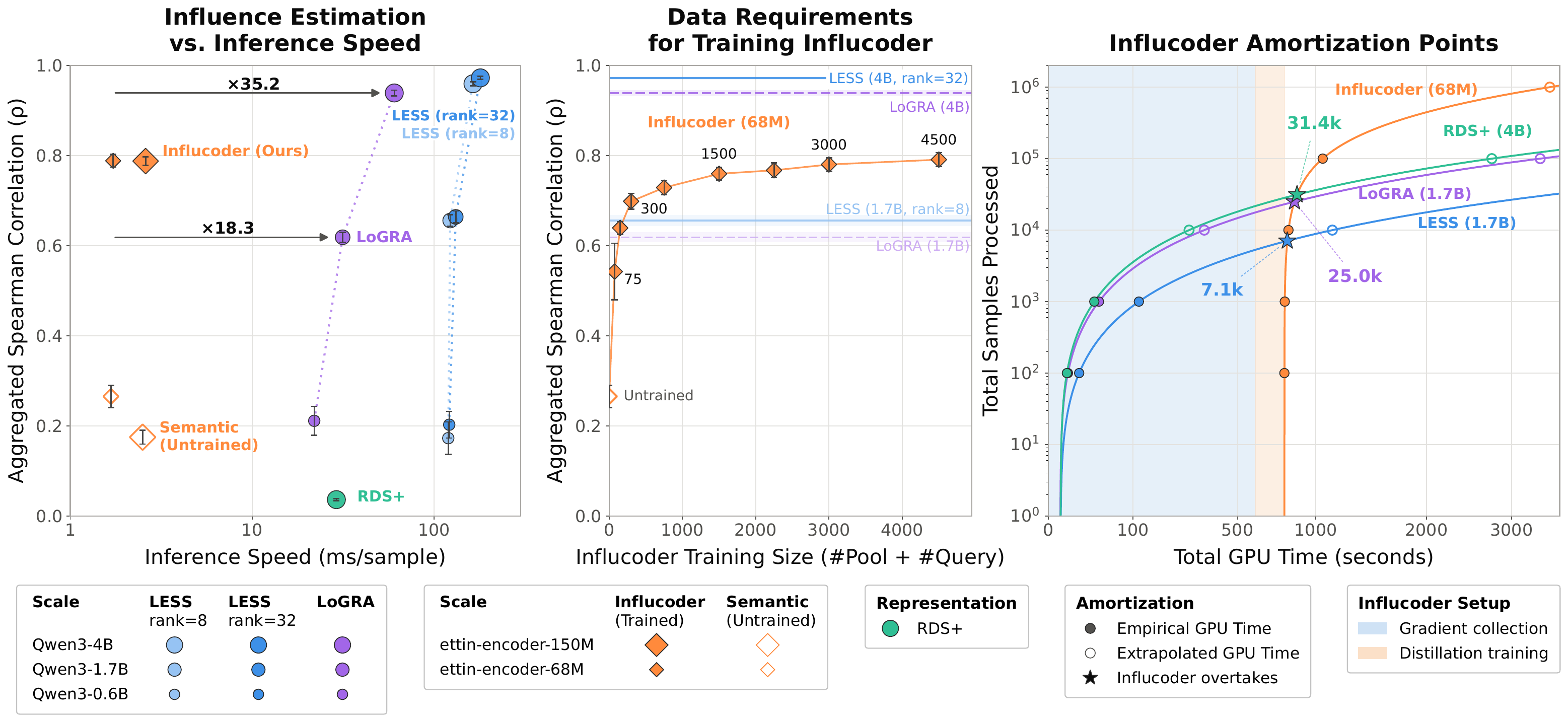}
  \caption{Influence ranking estimation from \textit{Dolci} to \textit{BBH} with
\textit{Qwen3-4B} as the target.
\textbf{Left:} Spearman correlation against inference speed. Influcoder outpaces gradient-based methods at $18.3\times$ and $35.2\times$ lower inference time
than LoGra on the 1.7B and 4B models. \textbf{Middle:} ranking quality against
distillation size, at a 1:2 query-to-pool ratio. \textbf{Right:}
total GPU time against samples processed. Stars mark where Influcoder
amortizes against the other methods.}
  \label{MainExp}
\end{figure*}

\subsection{Influence Ranking Estimation}

We attribute from a pool of Dolci \citep{olmo2025olmo3} instruction-tuning samples, from the \textit{tasksource} release \citep{sileo-2024-tasksource-large},
to a query set of BBH \citep{suzgun-etal-2023-challenging} reasoning tasks. We distill into encoders of the
\textit{ettin-encoder} suite \citep{weller2026seq} and compare
against LESS \citep{xia2024less}, LoGra \citep{choe2025what} and RDS+
\citep{ivisondata2025}, together with the untrained encoder, denoted
\textit{Semantic}. LESS and LoGra are run both on the target model, Qwen3-4B \citep{yang2025qwen3technicalreport}, and on the 1.7B and 0.6B variants as cheaper proxies. We exclude IProX
\citep{chen2026influencepreserving}, whose full-gradient influences track the
LoRA influences too poorly to serve as a comparison. For our ranking estimation metric we report the mean-aggregated Spearman. For ranking estimation, we report the mean-aggregated Spearman, which is the Spearman correlation between ground-truth and predicted global influence, where each pool sample's global influence is its mean influence across all query samples. We evaluate all methods over 800 pool and 800 query samples and report the mean-aggregated Spearman correlation against inference time. 
We then measure how the amount of distillation data affects the estimation
quality, varying the number of training samples at a fixed 1:2 query-to-pool
ratio. All ranking results are averaged over three seeds. Finally, we run an amortization experiment, locating the number of
processed samples at which we recoup the setup costs. 
 
\subsection{Downstream Attribution Tasks}

We evaluate on two attribution tasks from \citet{2025datelm}, against
representation-based baselines (BM25, RepSim, and Semantic) and gradient-based
baselines (LESS~\citep{xia2024less}, GradSim, and DataInf~\citep{kwon2024datainf}).
RepSim~\citep{2025datelm} is a slight variation of RDS+~\citep{ivisondata2025}.
On both tasks we distill on data disjoint from the evaluation pool, so that no
sample seen during distillation appears at evaluation time. In \textit{XSTest-response} \citep{rottger-etal-2024-xstest, NEURIPS2024_0f69b4b9}, each method must surface the toxic examples of an
UltraChat pool \citep{ding-etal-2023-enhancing} by scoring it against a toxic
reference set. We train \textsc{Influcoder} on WildGuardMix query samples
\citep{NEURIPS2024_0f69b4b9}, as well as ToxicChat \citep{lin2023toxicchat}
and benign UltraChat pool samples. \citet{2025datelm} show that retraining with the
toxic examples removed yields less toxic model behavior, which motivates
performance on this task as correlating with reduced toxicity after retraining. In \textit{Counterfactual}, we
train on a disjoint split of counterfact-tracing \citep{meng2022locating} and
report Recall@50 and MRR. 
 
\section{Results}
 
\begin{figure}[h]
  \centering
  \includegraphics[width=0.88\columnwidth]{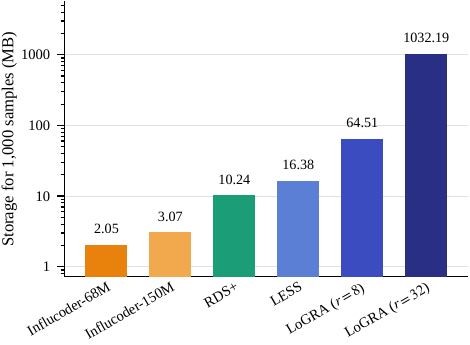}
  \caption{Storage footprint for each attribution method on 1K samples with \textit{Qwen3-4B} as the target decoder. LoGra rank quickly inflates storage requirements. 
  }
  \label{fig:storage}
\end{figure}

\paragraph{Ranking estimation}
Relative to Influcoder using ettin-encoder-68m, LoGra, which achieves the best inference speed to Spearman correlation performance of the baselines, is 18.3$\times$ slower on Qwen3-1.7B, and 35.2$\times$ slower on the Qwen3-4B model (Figure~\ref{MainExp}). Interestingly, the 68m encoder performs on par with the 150m encoder at reduced cost. RDS+ falls below every gradient-based configuration since influence rankings do not necessarily track with activation or semantic similarity. \textsc{Influcoder} also requires the least storage of all methods (Figure~\ref{fig:storage}). Each example is stored as a single encoder embedding, whose size depends only on the encoder and not on the target
decoder. Gradient-based baselines instead store projected gradients whose size
grows with the projection dimension, or rank as the LoGra configurations
show.  
 
\paragraph{Distillation cost}

The minimum amount of data sufficient to surpass all of the proxy models' influence ranking estimations is 300 samples, after which the quality of the estimations gradually increases when scaling up the samples. On the amortization, the setup required for our method becomes worthwhile compared to LESS on Qwen3-1.7B after 7.1k processed samples and LoGra after 25.0k samples. We note that these amortization points are given against the proxy configurations,
which our method already surpasses in ranking quality.

\paragraph{Downstream attribution}
We stay within 0.013 AUPRC of LESS on XSTest-response-Het
and within 0.044 on the Homogeneous variant (Table~\ref{tab:xstest_results}), and on Counterfact it comes
within 0.039 of Recall@50 against LESS (Table~\ref{tab:counterfactual_results}). Even once the one-time setup is
included in the total time cost, we are quicker in processing the datasets against every
gradient-based method on both tasks.

\begin{table}[ht]
    \centering
    \resizebox{\columnwidth}{!}{%
    \begin{tabular}{lcccc}
        \toprule
        & \multicolumn{2}{c}{XSTest-response-Het} & \multicolumn{2}{c}{XSTest-response-Hom} \\
        \cmidrule(lr){2-3} \cmidrule(lr){4-5}
        Method & AUPRC & Time (s) & AUPRC & Time (s) \\
        \midrule
        BM25       & 0.292 & -- & 0.295 & -- \\
        \textsc{Influcoder} & \underline{0.646}   & \underline{125.9}  & \underline{0.714} & \textbf{125.4} \\
        \quad \textit{+ setup} & -- & \textit{(+247.4)} & -- & \textit{(+350.8)} 
        \\
        Semantic   & 0.544 & \textbf{125.7} & 0.554 & \underline{126.0} \\
        RepSim     & 0.404 & 229.7 & 0.422 & 227.1 \\
        LESS       & \textbf{0.659} & 4801.4 & \textbf{0.758} & 4726.5 \\
        DataInf    & 0.152 & 5036.5 & 0.176 & 5079.4 \\
        GradSim    & 0.631 & 1372.9 & 0.689 & 1476.3 \\
        \bottomrule
    \end{tabular}%
    }
    \caption{Toxicity filtering on XSTest-response with \textit{Pythia-1B} \citep{biderman2023pythia} as
    the target model over two 10K sample UltraChat pools, each containing 66 toxic
    examples.}
    \label{tab:xstest_results}
\end{table}

\begin{table}[ht]
    \centering
    \resizebox{\columnwidth}{!}{%
    \begin{tabular}{lccc}
        \toprule
        & \multicolumn{3}{c}{Counterfactual Attribution} \\
        \cmidrule(lr){2-4}
        Method & Recall@50 & MRR & Time (s) \\
        \midrule
        BM25       & 0.349 & 0.583 & -- \\
        \textsc{Influcoder} & 0.460 & \underline{0.830} & \textbf{6.7} \\

        \quad \textit{+ setup} & -- & -- & \textit{\small(+462.2)} \\
        
        Semantic   & 0.279 & 0.753 & \underline{8.1} \\
        RepSim     & 0.376 & 0.791 & 66.9 \\
        LESS       & \textbf{0.499} & 0.767 & 2606.7 \\
        DataInf    & 0.472 & 0.765 & 6438.7 \\
        GradSim    & \underline{0.493} & \textbf{0.836} & 889.0 \\
        \bottomrule
    \end{tabular}%
    }
    \caption{Counterfactual attribution with \textit{Pythia-1B} as the target
    model. Each method retrieves, for each of the 66 queries, the training
    samples expressing the queried fact from a pool of 5{,}473 candidates.}
    \label{tab:counterfactual_results}
\end{table}

\section{Conclusion}
We presented \textsc{Influcoder}, which distills the LoRA-gradient influence
rankings of a target decoder into a compact encoder and reduces influence
estimation to a single encoder embedding pass. The distilled encoder recovers
the target's rankings at a 18$\times$ to 35$\times$ lower
inference time than LoGra on the proxy and target models respectively. Additionally, we show that our influence estimations retain enough fidelity for downstream applications in toxicity filtering and factual attribution. Our influence embeddings are compact, 512 dimensions for the smallest encoder, and independent of the target decoder's size. Future work could exploit this by indexing large pools for influence-based retrieval with approximate nearest-neighbour frameworks such as FAISS \citep{douze2024faiss}.

\section*{Limitations}

The performance of our method drops significantly on out-of-distribution data not seen during training. Consequently, plug-and-play methods such as LESS or LoGra are better suited to applications that must handle arbitrary data sources natively. Our method also requires a minimum amount of training data, which makes it ill-suited to settings where little or no training data is available. Furthermore, we do not compute exact influence: we restrict ourselves to LoRA-cosine influence and omit the inverse Hessian term needed to account for interactions between training samples. An open question is how much larger the target decoder can be relative to the encoder. Since influence is tied to how the model reasons about a problem, a decoder that is too large may exhibit dynamics too complex for the encoder to disentangle. Finally, given the limited generalization of our method, we have not tested whether it can be quickly re-fitted to new distributions.

\section*{Acknowledgements}
This work was supported by the French National Research Agency (ANR) through the ANR-24-CE23-4637 grant (Adada project). Generative AI tools were used for language editing, rewriting and coding tasks. Experiments presented in this paper were carried out using Grid5000, one of the precursor infrastructures of the SLICES-FR testbed, operated by a scientific interest group including 14 major French academic institutes and Universities (see \href{https://slices-fr.eu/}{https://slices-fr.eu/}).

%%%
% While we position ourselves on large-scale attribution. For smaller
% datasets, plug-in methods such as LESS and LoGRA are a better fit, as they
% require no training and apply to any pool and query set. Separately, our
% encoder learns the influence rankings of a single distribution and requires
% sufficient data from that distribution to do so, which makes tasks with few
% available queries less suitable. Furthermore, we estimate LoRA-cosine influences, which diverges from. Our evaluation covers one encoder family at modest scale. Whether the results
% transfer to other encoder families and larger encoders, to larger target models, and to further datasets remains untested.

\bibliography{custom}

\clearpage

\appendix

\section{Implementation Details}

\subsection{Hyperparameters}

Table~\ref{tab:settings-ranking} gives the hyperparameters and
settings for the ranking estimation experiment, while
Table~\ref{tab:settings-downstream} gives the parameters for the downstream evaluation experiments.

\begin{table}[h]
\centering
\begin{tabular}{lc}
\toprule
\textbf{Setting} & \textbf{Ranking est.} \\
\midrule
GPU                        & H200 NVL \\
Student encoder            & ettin 68m / 150m \\
Eval pool / query          & 800 / 800 \\
CountSketch proj.\ dim.    & 65{,}536 \\
Train pool size            & 3{,}000 \\
Train query size           & 1{,}500 \\
Batch size                 & 8 \\
Candidates per batch       & 16 \\
Learning rate              & 5e-5 \\
High-influence ratio        & 0.0 \\
Epochs                     & 8 \\
Alpha ($\alpha$)           & 0.5 \\
Weight decay               & 0.01 \\
\bottomrule
\end{tabular}
\caption{Hyperparameters and settings for the influence ranking
estimation experiment.}
\label{tab:settings-ranking}
\end{table}

\begin{table}[h]
\centering
\resizebox{\columnwidth}{!}{%
\begin{tabular}{lccc}
\toprule
& \multicolumn{3}{c}{Downstream} \\
\cmidrule(lr){2-4}
\textbf{Setting} & \textbf{Factual attr.} & \textbf{Toxicity (Hom.)} & \textbf{Toxicity (Het.)} \\
\midrule
GPU                        & A40 & A40 & A40 \\
Student encoder            & ettin 400m & ettin 400m & ettin 400m \\
Eval pool / query          & 5{,}473 / 66 & 10{,}066 / 10 & 10{,}187 / 10 \\
CountSketch proj.\ dim.    & 8{,}192 & 8{,}192 & 8{,}192 \\
Train pool size            & 2{,}000 & 1{,}300 & 1{,}300 \\
Train query size           & 900 & 100 & 100 \\
Batch size                 & 8 & 8 & 8 \\
Candidates per batch       & 16 & 16 & 16 \\
Learning rate              & 2e-5 & 2e-5 & 2e-5 \\
High-influence ratio        & 0.25 & 0.25 & 0.25 \\
Epochs                     & 8 & 8 & 8 \\
Alpha ($\alpha$)           & 0.5 & 0.5 & 0.5 \\
Weight decay               & 0.01 & 0.01 & 0.01 \\
\bottomrule
\end{tabular}%
}
\caption{Hyperparameters and settings for the downstream evaluation
experiments.}
\label{tab:settings-downstream}
\end{table}

\subsection{Algorithmic Complexity}

The complexity of training \textsc{Influcoder} pertains to two parts. The Gradient Collection step is generally the most time-intensive step, as is detailed in Figure~\ref{MainExp}.
\paragraph{Gradient Collection}
Gradient collection runs in $O(Q+P)$ time since we need to collect the gradients for $P$ pool samples and $Q$ query samples, each of which takes constant time.
\paragraph{Influcoder Training}
For the actual training of the encoder the complexity under one epoch is $O(Q \times P_{Batch})$, where $Q$ is the total number of query samples. Then, $P_{Batch}$ is the number of pool samples per batch, which for example equals 16 for the experiment in Table \ref{tab:xstest_results}. This is because on every batch we only sample a small subset of pool samples rather than visiting all possible combinations.

\section{Ablations}

\subsection{Ground-Truth Projection Validation}
\label{Appendix-A1}
Our labels are themselves projected, so it is worth knowing what that costs.
Table~\ref{tab:proj-validation} gives mean-aggregated Spearman correlations
between dense ground-truth gradients and projections of several sizes.

\begin{table}[htbp]
    \centering
    \begin{tabular}{ccc}
        \toprule
        Proj. Dim & Per-Anchor ($\rho$) & Aggregated ($\rho$) \\
        \midrule
        8{,}192  & 0.9549 & 0.9803 \\
        16{,}384 & 0.9649 & 0.9886 \\
        32{,}768 & 0.9700 & 0.9918 \\
        65{,}536 & 0.9737 & 0.9914 \\
        \bottomrule
    \end{tabular}
    \caption{Ranking retained across projection dimensions (Qwen3-1.7B). The
    pattern is consistent across model families, with larger models needing
    larger projections for the same estimation quality.}
    \label{tab:proj-validation}
\end{table}

\subsection{Alpha Value Ablation}

\begin{figure}[h]
  \centering
  \includegraphics[width=0.88\columnwidth]{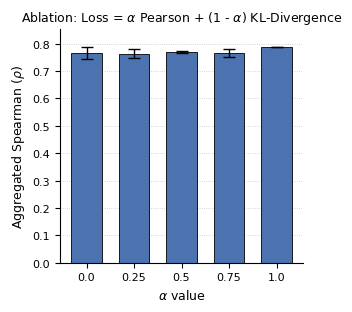}
  \caption{Different values of $\alpha$ still produce loss functions that are stable and achieve similar performance, albeit, the $\alpha$ value can be tuned to achieve small performance gains.
  }
  \label{fig:alphaAblation}
\end{figure}

\section{Generalization}

\begin{figure*}[t]
  \includegraphics[width=0.90\linewidth]{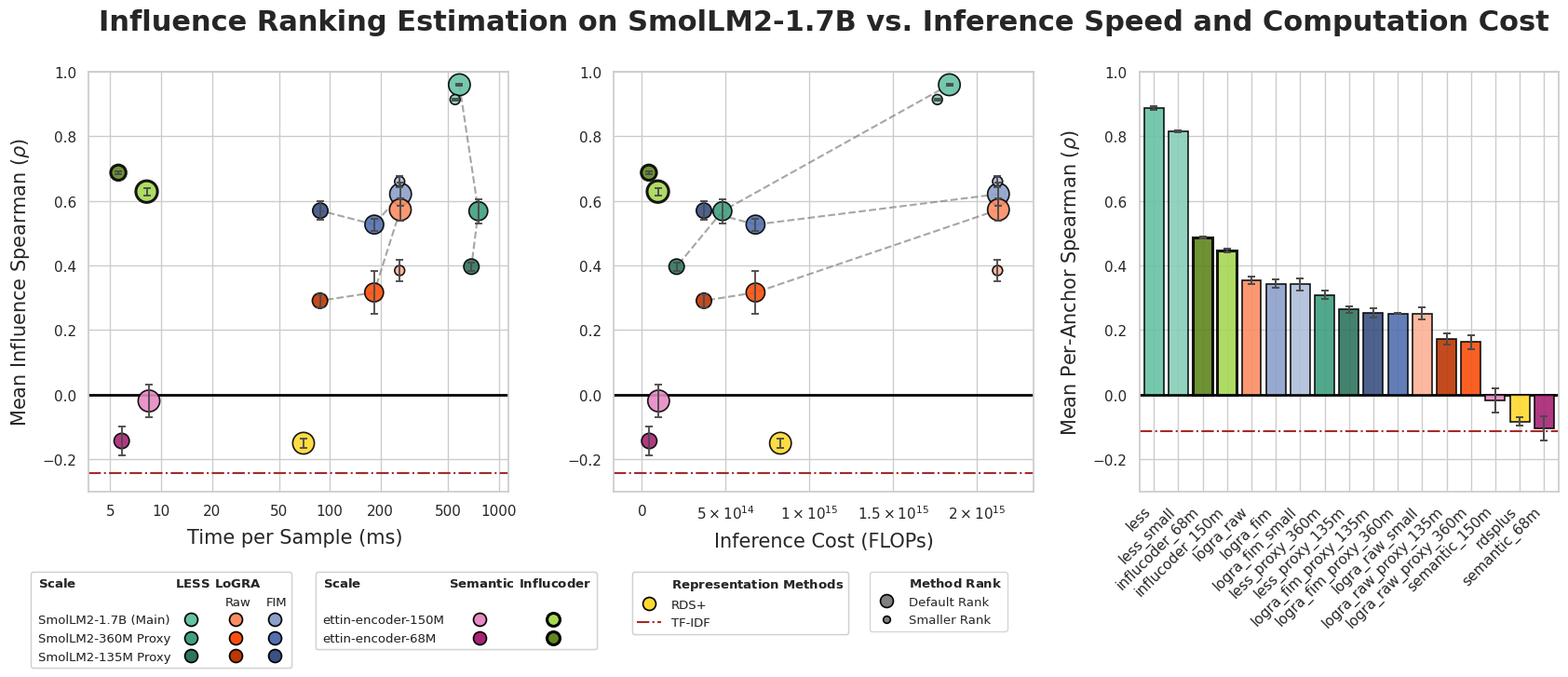} \hfill
  \caption {We repeat the ranking experiment with SmolLM2-1.7B~\citep{allal2025smollm}
as the target model, using Dolly~\citep{DatabricksBlog2023DollyV2} as the pool
and BBH~\citep{suzgun-etal-2023-challenging} as queries.}
\label{app:repro}
\end{figure*}

\subsection{Additional Dataset and Decoder Family Reproduction}
\label{app:reprosec}

Our ranking experiment results remain consistent across changes in the pool dataset from \textit{Dolci} to \textit{Dolly} as well as a different model family, \textit{SmolLM2} for the target decoder (Figure~\ref{app:repro}).

\subsection{Out-of-Distribution Evaluation}
\label{app:ood}
We evaluate an \textsc{Influcoder} distilled on Dolly pool samples
(SmolLM2-1.7B target, BBH queries, the setup is similar to Appendix~\ref{app:reprosec})
on a pool drawn from Fineweb~\citep{penedo2024the}, which it did not see during distillation.
Table~\ref{tab:ood} reports the aggregated Spearman correlation on both pools.

\begin{table}[h]
\centering
\begin{tabular}{lc}
\toprule
Eval pool & Aggregated Spearman ($\rho$) \\
\midrule
Dolly  & 0.60 \\
Fineweb & 0.06 \\
\bottomrule
\end{tabular}
\caption{Ranking quality in and out of the distillation distribution.}
\label{tab:ood}
\end{table}

\end{document}